\documentclass[letterpaper]{article} 
\usepackage{aaai25}  
\usepackage{times}  
\usepackage{helvet}  
\usepackage{courier}  
\usepackage[hyphens]{url}  
\usepackage{graphicx} 
\usepackage{natbib}  
\usepackage{caption} 
\usepackage{algorithm}
\usepackage{algorithmic}
\usepackage[dvipsnames,table]{xcolor}

\usepackage{amsthm}
\usepackage{amsmath, bm}
\usepackage{amsfonts}
\DeclareMathOperator*{\argmin}{arg\,min}

\usepackage{upgreek}
\usepackage{svg}
\usepackage{booktabs}
\usepackage{multirow}
\usepackage{subfig}

\usepackage{fontawesome5}

\usepackage{enumitem}
\usepackage{subcaption}

\newcommand{\first}[1]{\cellcolor{Orchid!8}\textcolor{Plum!30!black}{\textbf{#1}}}

\newcommand{\textpurple}[1]{\colorbox{Orchid!8}{\textcolor{Plum!30!black}{#1}}}

\usepackage{newfloat}
\usepackage{listings}
\DeclareCaptionStyle{ruled}{labelfont=normalfont,labelsep=colon,strut=off} 
\floatstyle{ruled}
\newfloat{listing}{tb}{lst}{}
\floatname{listing}{Listing}
\title{Real-time Calibration Model for Low-cost Sensor in Fine-grained Time series}
\author {
    Seokho Ahn\textsuperscript{\rm 1},
    Hyungjin Kim\textsuperscript{\rm 1},
    Sungbok Shin\textsuperscript{\rm 2}\thanks{The work was done while the author was affiliated with the University of Maryland, College Park.},
    Young-Duk Seo\textsuperscript{\rm 1}\thanks{Corresponding author.}
}
\affiliations {
    \textsuperscript{\rm 1}Department of Electrical and Computer Engineering, Inha University, Incheon 22212, South Korea\\
    \textsuperscript{\rm 2}Team Aviz, Inria, Universit\'e Paris-Saclay, Saclay, France \\
    sokho0514@inha.edu, flslzk@inha.edu, sungbok.shin@inria.fr, mysid88@inha.ac.kr
}

\begin{document}

\maketitle

\begin{abstract}
Precise measurements from sensors are crucial, but data is usually collected from low-cost, low-tech systems, which are often inaccurate. 
Thus, they require further calibrations. 
To that end, we first identify three requirements for effective calibration under practical low-tech sensor conditions.
Based on the requirements, we develop a model called \textbf{\textsc{Tesla}}, \textbf{T}ransformer for \textbf{e}ffective \textbf{s}ensor calibration utilizing \textbf{l}ogarithmic-binned \textbf{a}ttention. 
\textsc{Tesla} uses a high-performance deep learning model, Transformers, to calibrate and capture non-linear components.
At its core, it employs \textit{logarithmic binning} to minimize attention complexity. 
\textsc{Tesla} achieves consistent real-time calibration, even with longer sequences and finer-grained time series in hardware-constrained systems. 
Experiments show that \textsc{Tesla} outperforms existing novel deep learning and newly crafted linear models in accuracy, calibration speed, and energy efficiency. 
\end{abstract}

%

\section{Introduction}
\label{sec:01_intro}

Externalizing data latent in our world~\cite{elmqvist23anytime} can increase our awareness of previously unknown situations. 
For example, air pollution is a pervasive issue often overlooked due to its invisibility, and increased awareness can help us avoid potential dangers. 
Sensors from various IoT devices are used to capture real-time information from our surroundings.
To take well-informed, suitable actions about our environments, it is integral to obtain accurate data from sensors. 
But oftentimes, these sensors exist in the form of low-spec computing devices, and these devices are sometimes not accurate \cite{ray2022tinyml}.
To address this matter, there have been attempts to improve the quality of data coming from these low-tech devices using na\"ive linear- and machine learning-based calibration methods \cite{concas2021machinelearningcalibration, aula2022machinelearningcalibration,villanueva2023mutlisensor}.

We identify three key requirements that must be addressed to develop an effective, and practical calibration model: (1) handling fine-grained sensor type, (2) ensuring consistency in real-time calibration, and (3) accommodating hardware constraints (Section \ref{sec_problem}).
Recent research has utilized time series prediction for calibrating low-cost sensors, but practical IoT systems face limitations in their effective use.
Researchers have adopted time series prediction methods for low-cost sensor calibration~\cite{zhang23stcm,ahn24sendal,narayana24sensbert} as both rely on similar time series regression models~\cite{narayana24sensbert,toner2024analysislineartimeseries}.
This, together with deep learning and linear-based approaches, has led to improved calibration accuracy, but there is a tradeoff --- an improvement in one factor leads to a setback in others~\cite{zhang23stcm,ahn24sendal,narayana24sensbert}.
Addressing all three of the factors mentioned above in tandem is not a trivial issue.

To bridge this research gap, we propose \textsc{Tesla} (\textbf{T}ransformer for \textbf{e}ffective \textbf{s}ensor calibration utilizing \textbf{l}ogarithmic-binned \textbf{a}ttention) to address all of these challenges for IoT systems. 
\textsc{Tesla} reduces the attention bottleneck in transformers through logarithmic binning (Section \ref{sec:04_method_binning}), which couples more past tokens while retaining recent ones in a log scale. 
\textsc{Tesla} also employs multi-view embedding (Section \ref{sec:04_method_embedding}) and feature-wise aggregation (Section \ref{sec:04_method_output}) to preserve both local and global time series patterns, using only a single sensor.
\textsc{Tesla} operates as quickly as linear-based models and can be integrated into low-tech systems, offering higher accuracy than deep learning-based models to balance the tradeoffs. 
Experimental results demonstrate that \textsc{Tesla} successfully achieves this balance, efficiently managing calibration speed, energy usage, and accuracy.

To sum up, our contributions are: 

\begin{itemize}[leftmargin=1em]
    \item We identify three challenges that have to be addressed to achieve accuracy on par with high-quality sensors within practical IoT systems.
    \item We propose a model called \textsc{Tesla} to address all of these challenges for low-cost sensor calibration.
    \item Experiments with real-world benchmarks show that \textsc{Tesla} outperforms the baseline deep learning and newly designed linear models in most cases.
\end{itemize}


\section{Related Works}
\label{sec:02-background}

We present the related work from two perspectives: low-cost sensor calibration and short-term time series prediction.

\subsection{Low-cost Sensor Calibration}
Existing works on low-cost sensors tackle specific challenges, but their efficacy is limited as performance improvements come with setbacks.
We present these calibration trade-offs from two groups: (1) Machine learning-based and statistical models, and (2) deep learning-based models.

Various machine learning-based and statistical approaches such as linear regression, random forests, and ARIMA-based models have been used to calibrate low-cost sensors \cite{concas2021machinelearningcalibration, aula2022machinelearningcalibration,villanueva2023mutlisensor}. 
In particular, linear regression has been widely used due to its simplicity, but it has often struggled to capture nonlinear trends in time series \cite{concas2021machinelearningcalibration, patton2022nonlinear}.

Recent advances in deep learning have introduced various data-driven calibration methods, such as long short-term memory (LSTM) \cite{hochreiter1997lstm}-based \cite{nath2021statisticalanddeeplearning, ahn24sendal, apostolopoulos2023lstm}, convolutional neural networks (CNNs) \cite{krizhevsky2012cnn}-based \cite{ali2023deepcalib}, or Transformer \cite{vaswani2017attentionneed}-based methods \cite{narayana24sensbert, ahn24sendal}. 
However, training these models usually demands significant computing power, posing challenges for practical use in IoT-controlled systems~\cite{nath2021statisticalanddeeplearning}.
For example, the deployment of Transformers in everyday home devices is limited due to their high resource demands.
This highlights a significant gap in the practical application of advanced models in resource-constrained environments.

\subsubsection{Our contributions. }
\textsc{Tesla} meets all three requirements: it achieves high accuracy with deep learning methods and fastens inference speeds by partially implementing linear methods, whereas existing works mainly address one/two of the three issues, at the expense of other requirements.

\subsection{Short-term Time Series Forecasting}
Even though they handle similar tasks, time series forecasting models can not be used directly for calibration, as they do not meet the needs of practical calibration models.
Here, we describe two types of forecasters: (1) Transformer-based and (2) linear-based forecasters. 

The effectiveness of Transformers has been questioned by the advent of linear forecasters \cite{zeng2023dlinear}, but recent models like PatchTST \cite{nie2023timeseriesworth64} and iTransformer \cite{liu2024itransformer} have demonstrated that Transformers can effectively capture the time series patterns when timely used. 
Despite these advances, the complexity of Transformer-based models still poses challenges for practical IoT systems. 
Although iTransformer \cite{liu2024itransformer} is the most practical option, calibrating it with a single sensor is ineffective as it is specifically designed for multivariate time series. 

Linear forecasters are mainly advantageous for their computational efficiency and low energy use.
This makes them ideal for IoT systems.
However, these models have not yet addressed two critical issues when applied as calibration models: 
(i) Data collected from various sensors may be inaccurate due to its non-linear characteristics, which are more complex than in time series prediction \cite{concas2021machinelearningcalibration, patton2022nonlinear};
(ii) Their ability to handle large datasets in practical applications remains unproven due to the small size of the parameters. 
These limitations must be resolved for successful implementation in practical IoT systems.

\subsubsection{Our contributions. }
In contrast to many Transformer-based models, \textsc{Tesla} is tailored for fast and efficient calibration of a single time series, typical for most low-cost sensors. 
Furthermore, \textsc{Tesla} merges the strengths of both forecasters to achieve high speed (comparable to linear models) and high accuracy with limited resources.


\section{Preliminaries \label{sec_problem}}
\label{sec:03_prelim}

We define fine-grained time series and sensor calibration model.
We then discuss additional requirements when applying sensor calibration models to real-world scenarios.

\subsection{Fine-grained Time Series \label{sec_problem1}}

We first define the concept of fine-grained time series.

\smallskip
\theoremstyle{definition}
\newtheorem{define}{Definition}

\begin{define}[Fine-grained time series]
Consider a sensor \(\mathcal{X}\) as a function \(\mathcal{X}:T \rightarrow \mathbb{R}^+\), where \(T\) is a discrete set of measurement times. Then \(x_i=\mathcal{X}(t_i)\) represents the \(i_\textrm{th}\) sensor reading at time \(t_i\in T\). Given a window size \(N\), an \(i_\textrm{th}\) \textit{fine-grained time series} \(\mathcal{S}_i\) generated by sensor \(\mathcal{X}\) is a sequence \(\mathcal{S}_i=\left(x_{i-N+1}, \cdots, x_i\right)\) for \(i\geq N\), under the condition that the granularity \(\mu_\mathcal{X}=\min_{i\geq N}\left\{t_i-t_{i-1}\right\}\) of the sensor \(\mathcal{X}\) is sufficiently small. For notational convenience, we regard the time series \(\mathcal{S}_i\) as a fine-grained time series unless otherwise noted.~\footnote{Note that sensor sampling period can be delayed by several factors, such as sensor connection errors, invalid readings (\textit{e.g.}, null value), or valid readings that should be truncated (\textit{e.g.}, outliers). 
Hence, we define granularity as a minimum value of the sampling interval to ensure a consistent understanding of the granularity.}
\end{define}
\smallskip

After observing various time series benchmarks (e.g., ETT, traffic, weather, etc.), we have noted a growing demand for finer time series granularity, down to minute- or second-level intervals.
Calibration, in particular, demands finer granularity than most other forecasting tasks.
Air quality sensors equipped in home appliances must reflect dynamic, ad hoc environmental changes caused by several activities such as cooking, exhaust emissions, or smoke. 
For these applications, it is crucial to provide quick and accurate calibration results based on the recent sensor readings.
Our research thus focuses on these fine-grained time series.

\subsection{Sensor Calibration Model \label{sec_problem2}}
Here too, we start by defining sensor calibration model. 

\smallskip
\begin{define}[Sensor calibration model]
Consider two sensors performing the same task \(\mathcal{X}\) and \(\mathcal{Y}\), where sensor \(\mathcal{X}\) is less accurate than \(\mathcal{Y}\). Then a \textit{sensor calibration model} \(\mathcal{F}_N(*|\Theta_{\textrm{opt}})\) for sensor \(\mathcal{X}\) is a deep learning model, where the learnable parameters \(\Theta_{\textrm{opt}}\) of the model \(\mathcal{F}_N\) is optimized by the following formula with an objective function \(\mathcal{L}\):
\begin{equation}\label{eq:parameter_update}
    \Theta_{\textrm{opt}}=\argmin_{\Theta \in \Omega}\sum_{i\in I_{\textrm{train}}}\mathcal{L}\left(\mathcal{F}_N(\mathcal{S}_i|\Theta), y_i\right)
\end{equation}
where \(\Omega\) is the total parameter space of the model \(\mathcal{F}_N\) and \(I_{\textrm{train}}\subseteq\mathbb{N}\) denotes the training index set.~\footnote{The goal of the sensor calibration model \(\mathcal{F}_N\) is to calibrate sensor reading \(x_i\) using the sequence \(\mathcal{S}_i\), minimizing the difference between the predicted value \(\hat{x}_i=\mathcal{F}_N(\mathcal{S}_i|\Theta)\) and the reference reading \(y_i=\mathcal{Y}(t_i)\).} 
\end{define}\smallskip

Calibration tasks are better served by referencing external time series rather than relying solely on data from low-cost sensors. 
This is primarily because the data distributions from low-cost sensors and reference sensors can differ significantly. 
Hence, generating labels from a low-cost sensor itself may restrict their potential accuracy. 
In such scenarios, employing a high-quality sensor as a reference, such as those used in air quality monitoring stations, provides more reliable calibration results. 

However, optimizing learnable parameters is insufficient when designing a sensor calibration model. 
IoT devices equipped with sensors often have insufficient hardware resources, which limits the application of high-computational calibration models. 
This necessitates the need to satisfy requirements that match the capabilities of IoT devices, which we describe in the next subsection.

\subsection{Model Hardware Requirements \label{sec_problem3}}

This section describes three key factors that are required to develop an effective consistent, real-time calibration model for practical IoT scenarios:

\begin{itemize}[leftmargin=1em]
    \item \textbf{Accuracy. } Accurate data reading from sensors is crucial. 
    Commonly used evaluation metrics include mean absolute error (MAE) and root mean squared error (RMSE) to measure this accuracy. 
    However, in practical IoT systems, real-time performance matters, and constraints on IoT hardware may prevent the deployment of high-performance models.
    
    \item \textbf{Latency. } For finer-grained time series, calibration must be conducted swiftly, within smaller intervals.
    We can evaluate this latency using speed-related metrics, such as inference speed, floating point operations (FLOPS), or throughput.
    It is not necessary to choose the fastest calibration model; instead, the ideal choice would be a model that achieves the optimal performance given the situation.
    
    \item \textbf{Hardware resources. } Considering hardware resources is crucial when determining the applicability of a calibration model. 
    This involves evaluating model parameters to estimate hardware fixed memory requirements, memory footprints to estimate variable memory for space complexity, and calculating FLOPS to determine time complexity. 
\end{itemize}

In summary, evaluating the superiority of a calibration model requires considering both its accuracy and the additional factors we discussed. 
The ideal calibration model should exhibit the high accuracy of a deep learning model while maintaining the high speed and low energy consumption of linear models. 
In the next section, we propose \textsc{Tesla}, which aims to achieve this ultimate goal and will be detailed further. 
Then, in Section \ref{sec:05-experiment-setup}, we carefully select evaluation metrics that account for all these factors.


\section{Sensor Calibration Model: \textsc{Tesla} \label{sec_model}}
\label{sec:04-method}

This section describes our Transformer architecture for time series calibration for practical IoT systems, named \textbf{\textsc{Tesla}} (\textbf{T}ransformer for \textbf{e}ffective \textbf{s}ensor calibration utilizing \textbf{l}ogarithmic-binned \textbf{a}ttention).

\begin{figure}[t]
\centering
\includegraphics[width=1\columnwidth]{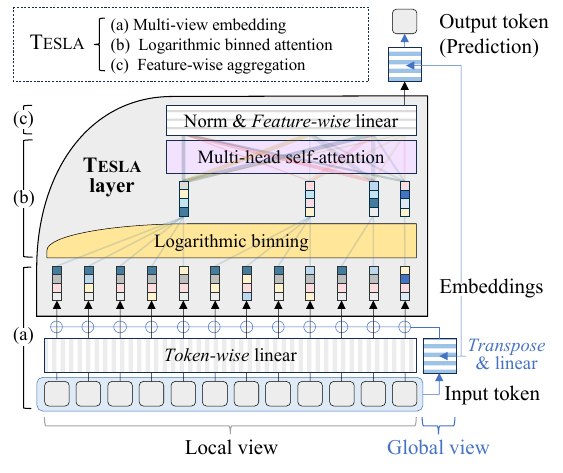}
\caption{Overview of \textsc{Tesla}, which consists of (a) multi-view embedding, (b) logarithmic binned attention, and (c) feature-wise aggregation in data process order. }
\label{fig:model}
\end{figure}

\subsection{Model Overview \label{sec:04_method_structure}}
Figure \ref{fig:model} shows the overall structure of \textsc{Tesla}. 
From the perspective of a black box model, \textsc{Tesla} takes the time series \(\mathcal{S}_i=\left(x_{i-N+1}, \cdots, x_i\right)\) as input and finally returns the calibration result \(\hat{x}_i\) as defined in the calibration model described in Section \ref{sec_problem2}. 
Our technical contribution lies in designing a novel architecture that meets the three hardware requirements mentioned in Section~\ref{sec_problem3}, while still maintaining high performance through the use of Transformers.

To begin with, \textsc{Tesla} uses multi-view embedding methods to input tokens to model both global and local features (Section \ref{sec:04_method_embedding}). 
This strategy effectively captures changes in fine-grained time series values while reducing the number of tokens. 
Second, we propose a novel method called logarithmic binning, aggregating a number of past tokens while preserving the detail of recent ones on a logarithmic scale (Section \ref{sec:04_method_binning}).
Logarithmic binning achieves significant \(\mathcal{O}(\log^2 N)\) self-attention complexity. 
It effectively alleviates the bottleneck problem of attention operations, making them applicable to IoT environments. 
Finally, we replace the token-wise feedforward network with a linear layer for feature-wise aggregation, which significantly reduces the number of learnable parameters and the computational power required (Section \ref{sec:04_method_output}). 
Detailed explanations are described in subsequent subsections.

\subsection{Multi-view Embedding\label{sec:04_method_embedding}}
We design an effective embedding method, even in the use of a single low-cost sensor. 
Our method is designed to consider multi-view data using simpler techniques, despite using only a univariate time series from a single sensor.
Relying solely on local token-wise embeddings, which are typical in Transformers, is ineffective for calibration because the receptive field is not large enough to effectively represent the time series. 
Hence, we add a global representation inspired by \citeauthor{liu2024itransformer} to preserve both local and global features.
This approach supplements implicit, local positional information within the global representation, removing the need for extra positional or temporal embeddings.

Formally, we first omit the index \(i\) and re-index the series as \(\mathcal{S}=\left(x_{1}, \cdots, x_N\right) \in \mathbb{R}^{1 \times N}\) (\textit{i.e.}, we shift the index by \(N - i\)) for notational convenience. 
Then the input representation of \textsc{Tesla} for a time series \(\mathcal{S}\) is defined as  \(\bm{\mathrm{E}}=\left(\bm{\mathrm{e}}_{1}, \cdots, \bm{\mathrm{e}}_N\right)^T\in \mathbb{R}^{N \times d}\), where:
\begin{equation}\label{logarithmic binning}
    \bm{\mathrm{e}}_{i} =
    x_i\bm{\mathrm{W}}_\text{local} + \mathcal{S}\bm{\mathrm{W}}_\text{global}
\end{equation}
is a \(d\)-dimensional column vector for \(i\in\left[1, N\right]\). \(\bm{\mathrm{W}}_{\text{local}}\in\mathbb{R}^{1\times d}\) and \(\bm{\mathrm{W}}_{\text{global}}\in\mathbb{R}^{N \times d}\) denote local and global learnable parameters, respectively.

\begin{figure}[t]
\centering
\includegraphics[width=1\columnwidth]{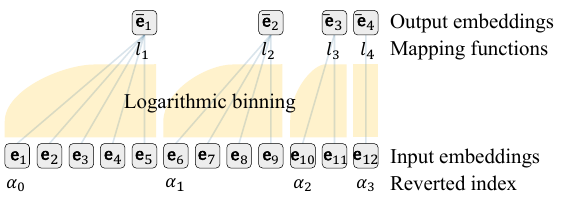}
\caption{Illustrative example of logarithmic binning in case \(N=12\). 
Then we have \(z=\left\lceil{\log_2 12}\right\rceil = 4\) with reverted indices \((\alpha_0, \alpha_1, \alpha_2, \alpha_3, \alpha_4)=(1, 6, 10, 12, 13)\). \(l_1, \cdots, l_4\) are mapping functions that satisfy Equation \ref{eq:lb}. }
\label{fig:lb}
\end{figure}

\subsection{Logarithmic Binned Attention \label{sec:04_method_binning}}

To alleviate the costly computation of Transformers, we deploy a core method called logarithmic binning. 
This reduces the number of tokens to approximately \(\log_2 N\).
Then we apply multi-head attention to these reduced tokens, enabling more efficient and rapid computations.

\subsubsection{Logarithmic binning. }
We first define \textit{logarithmic binning} as a collection of the mapping functions \(\left\{l_1, \cdots, l_z\right\}\) for \(z = \left\lceil{\log_2 N}\right\rceil\), where each function satisfies: 
\begin{equation}\label{eq:lb}
    l_j:\mathbb{R}^{d  \times (\alpha_j-\alpha_{j - 1})}\rightarrow\mathbb{R}^{d \times 1}
\end{equation}
\smallskip
where \(\alpha_j = \mathrm{max}\left\{1, N - 2^{z - j} + 3 \right\}\) for integer \(j \in \left[0, z\right]\). We provide an illustrative example of logarithmic binning including \(\alpha_j\)'s and \(l_j\)'s in Figure \ref{fig:lb} for comprehensive understanding. 
Intuitively, logarithmic binning is a strategy that keeps the latest token while a series of past tokens are grouped in reverse temporal order on a logarithmic scale. 
Using logarithmic binning, the sequence length is reduced from \(N\) to nearly \(\log_2 N\), enabling cost-efficient attention complexity \(\mathcal{O}(\log^2_2 N)=\mathcal{O}(\log^2 N)\).
Furthermore, the recent trend is emphasized, since a large part of recently inputted information is kept in the bin. 
Emphasizing these trends helps to better understand the recent sensor dynamics in long, finer-grained time series, leading to more accurate calibration results.

\noindent\paragraph{Logarithmic binned embedding. }
To reduce the number of tokens to reduce attention complexity, we then create an embedding by applying logarithmic binning to input embedding.
Using the previous collection of the mapping functions \(\left\{l_1, \cdots, l_z\right\}\), we define the logarithmic binned embedding of \(\bm{\mathrm{E}}\) as \(\overline{\bm{\mathrm{E}}}=(\overline{\mathbf{e}}_{1}, \cdots, \overline{\mathbf{e}}_z)^T\in \mathbb{R}^{z \times d}\), which is calculated by:
\begin{equation}\label{eq:lbe}
    \overline{\mathbf{e}}_{j}=l_j\left(\left(\mathbf{e}_{\alpha_{j-1}}, \cdots, \mathbf{e}_{\alpha_{j}-1}\right)\right).
\end{equation}
The most straightforward mapping function we can think of is the average function, which simply applies an average to each vector for binning. 
However, this method not only is incapable of providing learning opportunities but also loses information. 
Hence, we practically select these mapping functions as a set of learnable linear functions (act as learnable weighted average functions) to minimize the information loss while keeping it simple:
\begin{equation}\label{eq:lbe2}
    \overline{\mathbf{e}}_{j}=\left(\mathbf{e}_{\alpha_{j-1}}, \cdots, \mathbf{e}_{\alpha_{j}-1}\right)\bm{\mathrm{W}}_j
\end{equation}
where \(\bm{\mathrm{W}}_j \in \mathbb{R}^{(\alpha_j-\alpha_{j - 1}) \times 1}\) for \(j \in\left[1, z\right]\) are learnable parameters. 
This approach is capable of increasing the number of learnable parameters while using only approximately \(N\) additional parameters. 
This parameter efficiency can enable the model to operate effectively in practical IoT systems.

\subsubsection{Logarithmic binned attention. }
Finally, we perform a multi-head attention operation using only the reduced embedding of tokens to capture their interactions.
Using the logarithmic binned embedding \(\overline{\bm{\mathrm{E}}}\), we construct the query, key, and value as \(\overline{\bm{\mathrm{Q}}}=\overline{\bm{\mathrm{E}}}\bm{\mathrm{W}}_q\), \(\overline{\bm{\mathrm{K}}}=\overline{\bm{\mathrm{E}}}\bm{\mathrm{W}}_k\), and \(\overline{\bm{\mathrm{V}}}=\overline{\bm{\mathrm{E}}}\bm{\mathrm{W}}_v\) respectively, where \(\bm{\mathrm{W}}_q, \bm{\mathrm{W}}_k \), and \(\bm{\mathrm{W}}_v\in \mathbb{R}^{d \times d}\) are learnable parameters. 
Then the logarithmic binned self-attention is defined by:
\begin{equation}\label{eq:lba}
    \bm{\mathrm{Y}} = \mathrm{Softmax}\left(\frac{\overline{\bm{\mathrm{Q}}} \overline{\bm{\mathrm{K}}}^T}{\sqrt{d}}\right)\overline{\bm{\mathrm{V}}}.
\end{equation}
Note that \(\overline{\bm{\mathrm{Q}}} \overline{\bm{\mathrm{K}}}^T\) has a dimension \(z \times z\), resulting in the computational complexity of the logarithmic binned attention being \(\mathcal{O}(z^2)=\mathcal{O}(\log^2 N)\). 
We omitted multi-head concepts in the formulas since they are identical to those presented in a vanilla Transformer.

\begin{table*}[t]
\centering
\resizebox{1\textwidth}{!}{%
\setlength\tabcolsep{2.5pt}
\begin{tabular}{lrrrrrrrrrrrrrrrrrrrrrrrr}
\toprule
 & \multicolumn{8}{c}{\(\textbf{PM}_\textbf{10}\)} & \multicolumn{8}{c}{\(\textbf{PM}_\textbf{2.5}\)} & \multicolumn{8}{c}{\(\textbf{PM}_\textbf{1}\)} \\
 \cmidrule(l{0.5mm}r{0.5mm}){2-9} \cmidrule(l{0.5mm}r{0.5mm}){10-17} \cmidrule(l{0.5mm}r{0.5mm}){18-25}
 & \multicolumn{4}{c}{RMSE\scriptsize$\downarrow$} & \multicolumn{4}{c}{MAE\scriptsize$\downarrow$} & \multicolumn{4}{c}{RMSE\scriptsize$\downarrow$} & \multicolumn{4}{c}{MAE\scriptsize$\downarrow$} & \multicolumn{4}{c}{RMSE\scriptsize$\downarrow$} & \multicolumn{4}{c}{MAE\scriptsize$\downarrow$} \\
\textbf{Models} & \textbf{Ant.} & \textbf{Oslo} & \textbf{Zag.} & \textbf{Avg.} & \textbf{Ant.} & \textbf{Oslo} & \textbf{Zag.} & \textbf{Avg.} & \textbf{Ant.} & \textbf{Oslo} & \textbf{Zag.} & \textbf{Avg.} & \textbf{Ant.} & \textbf{Oslo} & \textbf{Zag.} & \textbf{Avg.} & \textbf{Ant.} & \textbf{Oslo} & \textbf{Zag.} & \textbf{Avg.} & \textbf{Ant.} & \textbf{Oslo} & \textbf{Zag.} & \textbf{Avg.} \\
 \cmidrule(l{0.5mm}r{0.5mm}){1-1} \cmidrule(l{0.5mm}r{0.5mm}){2-5}  \cmidrule(l{0.5mm}r{0.5mm}){6-9} \cmidrule(l{0.5mm}r{0.5mm}){10-13}
\cmidrule(l{0.5mm}r{0.5mm}){14-17} \cmidrule(l{0.5mm}r{0.5mm}){18-21}
\cmidrule(l{0.5mm}r{0.5mm}){22-25}
Raw & 23.05 & 23.64 & 15.41 & 20.70 & 20.57 & 10.75 & 8.04 & 13.12 & 16.41 & 8.46 & 5.98 &   10.28 & 15.15 & 13.81 & 5.97 & 11.64 & 13.21 &   6.48 & 5.58 & 8.42 & 9.59 & 5.13 & 4.57 & 6.43
\\
 \cmidrule(l{0.5mm}r{0.5mm}){1-1} \cmidrule(l{0.5mm}r{0.5mm}){2-5}  \cmidrule(l{0.5mm}r{0.5mm}){6-9} \cmidrule(l{0.5mm}r{0.5mm}){10-13}
\cmidrule(l{0.5mm}r{0.5mm}){14-17} \cmidrule(l{0.5mm}r{0.5mm}){18-21}
\cmidrule(l{0.5mm}r{0.5mm}){22-25}
Linear & 18.40 & 20.51 & 14.71 & 17.87 & 10.22 & 13.23 & 6.57 & 10.01 & 12.33 & 9.38 & 3.34 & 8.35 & 5.36 & 5.50 & 1.78 & 4.21 & 6.27 & 5.19 & 1.45 & 4.30 & 2.98 & 2.97 & 1.11 & 2.36 \\
NLinear & 20.78 & 22.12 & 15.45 & 19.45 & 14.38 & 13.89 & 6.81 & 11.69 & 14.80 & 9.91 & 5.74 & 10.15 & 10.14 & 6.20 & 3.66 & 6.67 & 8.89 & 6.68 & 3.62 & 6.40 & 5.98 & 4.36 & 2.69 & 4.34 \\
DLinear & 18.35 & 20.45 & 14.73 & 17.84 & 10.10 & 13.20 & 6.65 & 9.98 & 12.22 & 9.62 & 3.35 & 8.40 & 5.14 & 5.79 & 1.81 & 4.25 & 6.26 & 5.21 & 1.49 & 4.32 & 2.99 & 2.87 & 1.17 & 2.34 \\
 \cmidrule(l{0.5mm}r{0.5mm}){1-1} \cmidrule(l{0.5mm}r{0.5mm}){2-5}  \cmidrule(l{0.5mm}r{0.5mm}){6-9} \cmidrule(l{0.5mm}r{0.5mm}){10-13}
\cmidrule(l{0.5mm}r{0.5mm}){14-17} \cmidrule(l{0.5mm}r{0.5mm}){18-21}
\cmidrule(l{0.5mm}r{0.5mm}){22-25}
Transformer & 15.34 & 17.49 & 15.11 & 15.98 & 9.53 & 11.20 & 7.88 & 9.54 & 8.49 & 8.30 & 3.40 & 6.73 & 4.35 & 4.51 & 2.01 & 3.62 & \first{3.62} & 5.22 & 1.43 & \underline{3.42} & \first{1.83} & 2.51 & 1.14 & \underline{1.83} \\
Informer & 15.47 & \underline{16.36} & 14.86 & 15.56 & 9.48 & \underline{10.43} & 7.16 & 9.02 & \first{7.52} & 8.14 & 3.26 & \underline{6.31} & \underline{4.22} & 4.68 & \underline{1.66} & 3.52 & \underline{4.43} & 4.69 & \underline{1.34} & 3.49 & \underline{2.18} & 2.35 & \underline{1.00} & 1.84 \\
PatchTST & \underline{14.60} & 16.90 & 14.87 & \underline{15.46} & \first{9.21} & 10.80 & 7.43 & 9.15 & 7.66 & 8.29 & 3.33 & 6.43 & \first{4.18} & \underline{4.39} & 1.93 & \underline{3.50} & 6.14 & 4.86 & 2.57 & 4.52 &  3.70 & 3.13 & 2.38 & 3.07
\\ 
iTransformer & 16.54 & 16.40 & \underline{14.03} & 15.65 & 9.86 & 10.48 & \underline{6.38} & \underline{8.91} & 8.62 & \first{7.88} & \underline{3.23} & 6.58 & 4.58 & 4.58 & 1.83 & 3.66 & 5.10 & \underline{4.22} & 1.49 & 3.60 & 2.30 & \underline{2.24} & 1.32 & 1.95 \\
\textbf{\textsc{Tesla}} & \first{14.58} & \first{15.79} & \first{13.87} & \first{14.75} & \underline{9.36} & \first{10.10} & \first{6.26} & \first{8.57} & \underline{7.54} & \underline{7.90} & \first{3.03} & \first{6.16} & 4.34 & \first{3.96} & \first{1.38} & \first{3.23} & 4.78 & \first{3.63} & \first{1.14} & \first{3.19} & 2.34 & \first{2.01} & \first{0.80} & \first{1.71} \\
\bottomrule
\end{tabular}
}
\caption{Calibration accuracy in case \(N=360\) on three regions (\textbf{Ant.}, \textbf{Oslo}, and \textbf{Zag.}) and features (\(\textbf{PM}_\textbf{10}\), \(\textbf{PM}_\textbf{2.5}\), and \(\textbf{PM}_\textbf{1}\)). \textbf{Avg.} denotes the average performance across three regions. Raw indicates the difference between the low-cost sensor readings and the reference (without any calibration). The highest performance is marked in \textpurple{\first{bold}}, and the second highest in \underline{underlined}.}
\label{tb:result_main}
\end{table*}

\begin{table*}[t]
\centering

\resizebox{1\textwidth}{!}{%
\setlength\tabcolsep{7pt}
\begin{tabular}{lcccrrrrrrrrrc}
\toprule
\textbf{} & \multicolumn{3}{c}{\textbf{Architecture}} & \multicolumn{3}{c}{\(\textbf{PM}_\textbf{10}\)} & \multicolumn{3}{c}{\(\textbf{PM}_\textbf{2.5}\)} & \multicolumn{3}{c}{\(\textbf{PM}_\textbf{1}\)} \\ 
\cmidrule(l{0.5mm}r{0.5mm}){2-4} \cmidrule(l{0.5mm}r{0.5mm}){5-7} \cmidrule(l{0.5mm}r{0.5mm}){8-10} \cmidrule(l{0.5mm}r{0.5mm}){11-13}

\textbf{Models} & \textbf{Token} & \textbf{Embedding} & \textbf{Aggregator} & \textbf{Ant.} & \textbf{Oslo} & \textbf{Zag.}  & \textbf{Ant.} & \textbf{Oslo} & \textbf{Zag.} & \textbf{Ant.} & \textbf{Oslo} & \textbf{Zag.} & \textbf{Avg. gain} \\
\cmidrule(l{0.5mm}r{0.5mm}){1-1} \cmidrule(l{0.5mm}r{0.5mm}){2-4} \cmidrule(l{0.5mm}r{0.5mm}){5-7} \cmidrule(l{0.5mm}r{0.5mm}){8-10} \cmidrule(l{0.5mm}r{0.5mm}){11-13} \cmidrule(l{0.5mm}r{0.5mm}){14-14}

Transformer & Full & Local & Feed-forward & 15.34 & 17.49 & 15.11 & 8.49 & 8.30 & 3.41 &  \first{3.62} & 5.22 & 1.43 & -\\
\textsc{Tesla} & Binned (uniform) & Local & Feed-forward & 15.42 & 18.26 & 15.09 & 8.40 & \underline{8.02} & 3.36 & 4.41 & 4.96 & 1.35 & -1.20\%\\
\textsc{Tesla} & Binned (log-scaled) & Local & Feed-forward & 15.46 & 17.03 & 15.09 & 8.62 & 8.77 & 3.29 &  \underline{4.23} & 5.07 & 1.26 & -0.45\%\\
\textsc{Tesla} & Binned (log-scaled) & Local+Global & Feed-forward & \underline{15.03} & \underline{16.29} & \underline{14.54}  & \underline{7.94} & 8.16 & \underline{3.17} & 5.01 & \underline{4.70} & \underline{1.17} & \underline{+1.90\%}\\
\textbf{\textsc{Tesla} (Ours)} & Binned (log-scaled) & Local+Global & Linear & \first{14.58} & \first{15.79} & \first{13.87} & \first{7.54} & \first{7.90} & \first{3.03}  &  4.78 & \first{3.63} & \first{1.14} & \first{+6.88\%}\\ 
\bottomrule
\end{tabular}
}
\caption{Ablation studies in case \(N=360\) for Transformer and \textsc{Tesla} variants in terms of RMSE.  
\textbf{Avg. gain} represents the average increase in performance of the three regions compared to the Transformer.}
\label{tb:ablation}
\end{table*}

\subsection{Feature-wise Aggregation \label{sec:04_method_output}}
Networks that deploy Transformers typically implement token-wise processing with large hidden layers, which can pose computational overload on constrained hardware devices.
To address this, we get motivation from previous studies~\cite{zeng2023dlinear, liu2024itransformer}. 
We replace the token-wise feed-forward network, which incorporates a dual linear layer, with a single feature-wise linear layer. 
This modification more efficiently captures time series dynamics by repeating calculations only across the number of features.

Given an attention output \(\bm{\mathrm{Y}} \in \mathbb{R}^{z \times d}\), the final prediction \(\hat{x}\) is calculated by:
\begin{equation}\label{eq:aggregation}
    \hat{x} = (\mathrm{LayerNorm}(\bm{\mathrm{Y}})\bm{\mathrm{W}}_{\text{agg}1})^T\bm{\mathrm{W}}_\text{\text{agg}2}
\end{equation}
where \(\bm{\mathrm{W}}_{\text{agg}1}\in \mathbb{R}^{d \times 1}\) serves as a feature-wise aggregator, and \(\bm{\mathrm{W}}_{\text{agg}2}\in \mathbb{R}^{d \times 1}\) is used to transform the final token embedding into the scala prediction.


\section{Experiment Settings}
\label{sec:05-experiment}
\label{sec:05-experiment-setup}

Here, we first explain the datasets and training setup.
Then, we describe our baselines, evaluation metrics, and implementation details.

\subsubsection{Datasets. \label{sec:05-experiment-dataset}}
Our experiment is based on a large-scale dataset suitable for calibration task~\cite{van2023senseurcity}. 
This dataset includes sensor data collected from numerous sensors across three regions—Antwerp (\textbf{Ant.}), Oslo (\textbf{Oslo}), and Zagreb (\textbf{Zag.}), with three individual features---\(\textbf{PM}_\textbf{10}\), \(\textbf{PM}_\textbf{2.5}\), and \(\textbf{PM}_\textbf{1}\). 
Detailed descriptions of the dataset used in our experiment are shown in Appendix \ref{sec:appendix_dataset}.

\begin{figure*}[t]
\centering
\parbox{0.51\textwidth}{
\captionsetup[subfigure]{justification=raggedright, singlelinecheck=true, margin={0.5em, 0em}, indention=1.3em}
\vspace{-0.5em}
\subfloat[\textsc{Tesla} optimizes both RMSE and memory usage.]{
    \centering
    \includegraphics[width=0.217\textwidth]{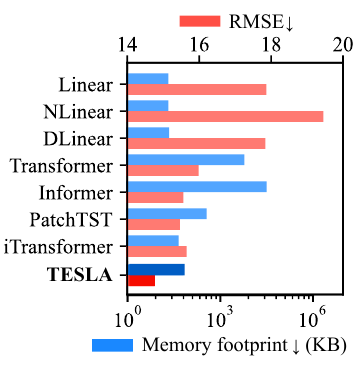}
    }
  \hfill
\captionsetup[subfigure]{justification=raggedright, singlelinecheck=true, margin={1.5em, 0em}, indention=1.3em}
\subfloat[\textsc{Tesla} balances RMSE, parameters, and FLOPS.]{
    \centering
    \includegraphics[width=0.24\textwidth]{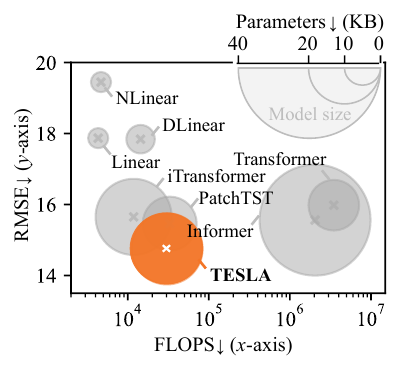}
    }
  \hfill
    \caption{Comparison of model efficiency for average \(\textbf{PM}_\textbf{10}\) performance across three regions in case \(N=360\). (a) Comparison of RMSE and memory footprint across models. (b) Scatter map of FLOPS, RMSE, and parameters (diameter of the circles).}
    \label{fig:exp_efficiency}
}
\hfill
\parbox{0.44\textwidth}{
\captionsetup[subfigure]{justification=raggedright, singlelinecheck=true, margin={0em, 0em}, indention=1.3em}
\vspace{-0.03em}
\subfloat[Performance rises with scaling but later declines.]{
    \centering
    \includegraphics[width=0.201\textwidth]{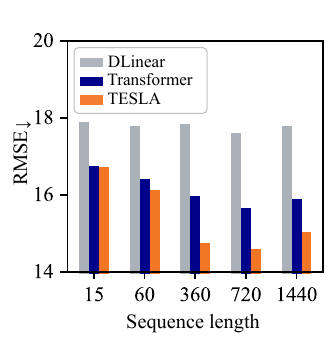}
    }
  \hfill
\captionsetup[subfigure]{justification=raggedright, singlelinecheck=true, margin={0.2em, 0em}, indention=1.3em}
\subfloat[\textsc{Tesla} scales efficiently, resembling linear model.]{
    \centering
    \includegraphics[width=0.205\textwidth]{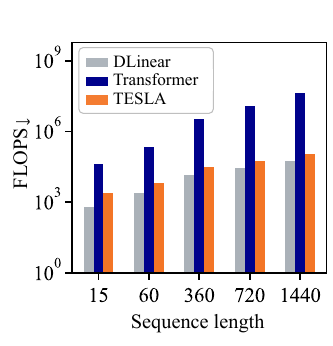}
    }
  \hfill
    \caption{Comparison of RMSE and FLOPS for average \(\textbf{PM}_\textbf{10}\) performance across three regions, evaluated at various time spans: 15 minutes, 1 hour, 6 hours, 12 hours, and 24 hours.}
    \label{fig:exp_window}
}
\end{figure*}

\begin{figure*}[t]
\centering
    \subfloat[Actual calibration results concentration over time.]{
    \centering
    \includegraphics[width=0.54\textwidth]{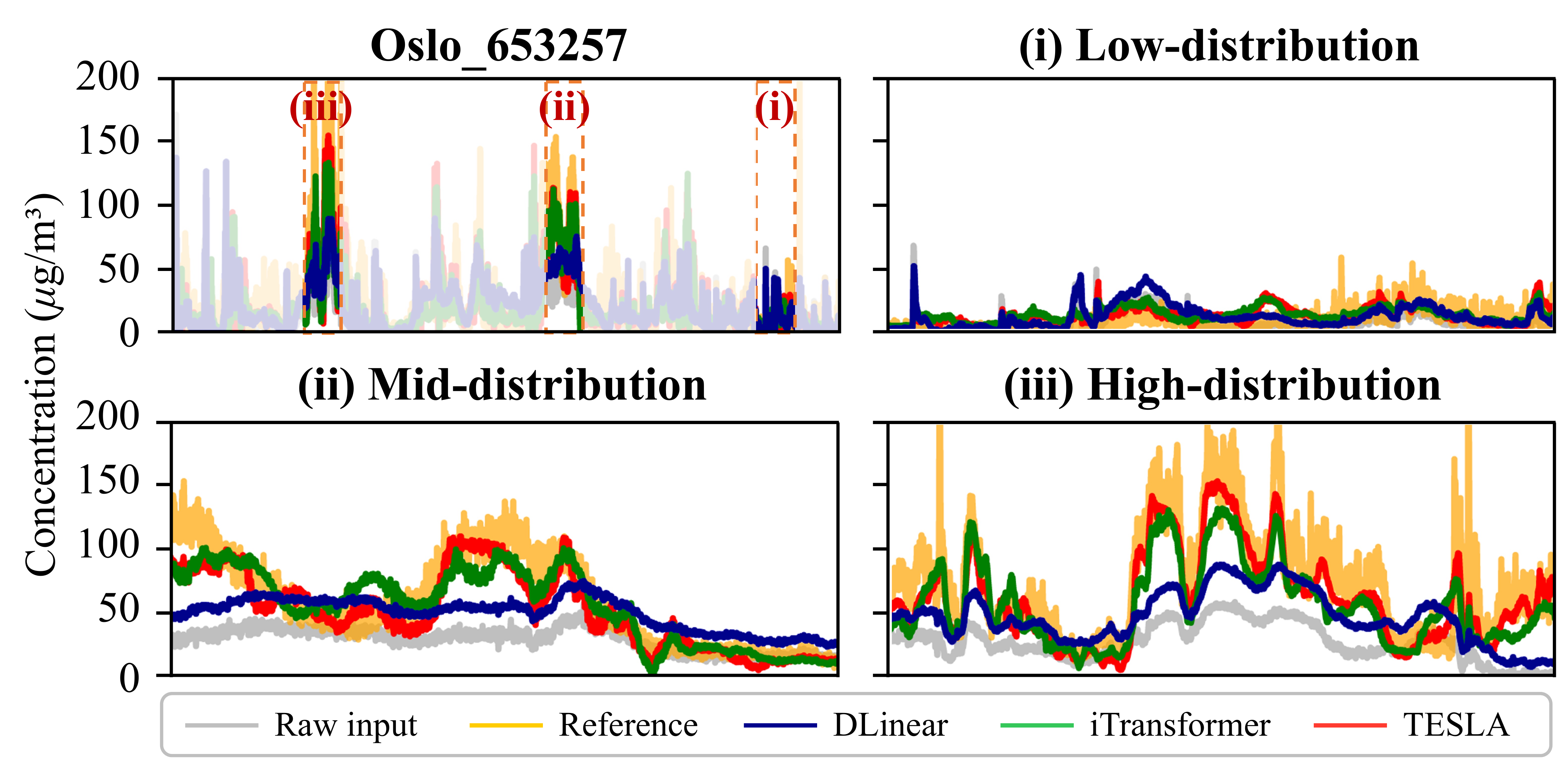}
    }
  \hfill
  \subfloat[Numerical comparisons for each distribution.]{%
    \centering
    \setlength\tabcolsep{4pt}
    \resizebox{0.44\textwidth}{!}{%
        \begin{tabular}[b]{lrrcrrc}
        \toprule
        & \multicolumn{3}{c}{\textbf{Oslo} \(\textbf{(PM}_\textbf{10}\textbf{)}\)} & \multicolumn{3}{c}{\textbf{(i) Low-distribution}} \\
        \textbf{Models} & RMSE & MAE & \textbf{Gain} & RMSE & MAE  & \textbf{Gain}\\
        \cmidrule(l{0.5mm}r{0.5mm}){1-1} \cmidrule(l{0.5mm}r{0.5mm}){2-4} \cmidrule(l{0.5mm}r{0.5mm}){5-7} 
        Raw & 23.64 & 10.75 & - & 8.78 & 6.09 & -  \\
        DLinear & 20.45 & 13.20 & -  & 9.64 & 6.59 & -  \\
        iTransformer & \underline{16.40} & \underline{10.48}  & -  & \underline{8.38} & \underline{6.81}  & -  \\
        \textbf{\textsc{Tesla}}  & \first{15.79} & \first{10.10} & \first{+3.81\%} & \first{8.26} & \first{6.48} & \first{+3.27\%} \\
        \midrule
        \\
        & \multicolumn{3}{c}{\textbf{(ii) Mid-distribution}}  & \multicolumn{3}{c}{\textbf{(iii) High-distribution}} \\
        \textbf{Models} & RMSE & MAE  & \textbf{Gain} & RMSE & MAE  & \textbf{Gain} \\
        \cmidrule(l{0.5mm}r{0.5mm}){1-1} \cmidrule(l{0.5mm}r{0.5mm}){2-4} \cmidrule(l{0.5mm}r{0.5mm}){5-7}  
        Raw & 42.17 & 30.08 & - & 57.66 & 47.67 & - \\
        DLinear & 30.55 & 23.60  & - & 42.22 & 32.67  & - \\
        iTransformer & \first{16.52} & \first{11.95}  & - & \underline{28.67} &  \underline{22.04}  & - \\
        \textbf{\textsc{Tesla}} &  \underline{16.58} &  \underline{12.20} &  \underline{-1.21\%} & \first{21.78} & \first{15.96}  & \first{+34.86\%} \\
        \bottomrule
        \end{tabular}
    }
  }
  \caption{Case study of actual \(\textbf{PM}_\textbf{10}\) calibration results with different calibration models for sensor \textbf{`Oslo\_643217'}, comparing results across three windows with different distributions (\textbf{low-}, \textbf{mid-}, and \textbf{high-distribution}). 
  \textbf{Gain} represents the average increase in performance in terms of RMSE and MAE compared to iTransformer.}
  \label{fig:exp_case}
\end{figure*}

\subsubsection{Evaluation setup. }

Our evaluation process assumes that we use multiple same types of sensors for training in a given space. 
Each sensors are uniquely identified by the name. 
We organize the sensors in alphabetical order: the second-to-last sensor is set as the validation set, and the last sensor as the test set. 
All remaining sensors are used for training. 
This configuration is repeated across different regions \textbf{Ant.}, \textbf{Oslo}, and \textbf{Zag.} and features \(\textbf{PM}_\textbf{10}\), \(\textbf{PM}_\textbf{2.5}\), and \(\textbf{PM}_\textbf{1}\).
A detailed evaluation setup is provided in Appendix \ref{sec:appendix_setup}.

\subsubsection{Baselines. }
We carefully select widely used state-of-the-art techniques for various time series tasks to compare with our proposed method, \textsc{Tesla}. 
For linear methods, we choose Linear \cite{freedman2009statistical}, NLinear \cite{zeng2023dlinear}, and DLinear \cite{zeng2023dlinear}, known for their fast inference speeds and outstanding performance in recent time series forecasting. 
For deep learning methods, we select Transformer \cite{vaswani2017attentionneed}, Informer \cite{zhou2021informer}, PatchTST \cite{nie2023timeseriesworth64}, and iTransformer \cite{liu2024itransformer}, known for their high performance but are less recognized in IoT environments. 
We give detailed explanations for each baseline in Appendix \ref{sec:appendix_baseline}.

\subsubsection{Evaluation metrics. }
The factors mentioned in Section~\ref{sec_problem3} must be considered when applying the calibration model to real-world scenarios. 
To comprehensively evaluate the model, we categorize the evaluation metrics into two groups: effectiveness and efficiency metrics.

For effectiveness, we measure accuracy using two widely adopted metrics in time series forecasting and calibration tasks: Root mean square error (RMSE) and mean absolute error (MAE).
For efficiency, we assess computational and resource requirements to ensure feasibility in practical IoT applications with low-cost sensors. 
Key metrics include floating point operations (FLOPS), memory footprint, and the number of model parameters.
Also, at the microcontroller level, inference speed and Flatbuffer size are considered to assess real-time applicability.

\subsubsection{Implementation. }
All experiments were conducted using a machine equipped with an AMD EPYC 7763 and an NVIDIA RTX A6000 Ada with TensorFlow 2.14, which supports conversion to TensorFlow Lite for Microcontrollers. 
All models were trained using a batch size of 32 and the Adam optimizer for 10 epochs with mean squared error objective function, following configurations commonly found in existing time series forecasting \cite{liu2024itransformer}. 
For evaluation on microcontrollers, the trained models were converted to FlatBuffers and deployed on an Arduino Nano 33 BLE Sense for evaluation.

\section{Experimental Results}
\label{sec_result}
This section summarizes the results of our work. 

\paragraph{Accuracy. } 
We first compared calibration accuracy. 
The results are shown in Table~\ref{tb:result_main}.
To sum up, overall, \textsc{Tesla} outperformed existing baselines in terms of RMSE and MAE.
Transformers outperformed linear models. This contrasts with previous observations in time series forecasting.
Among Transformer-based models, PatchTST and iTransformer demonstrated notable effectiveness in token compression for calibration tasks.
NLinear demonstrated the least accuracy among all linear-based models, with no significant difference in accuracy between Linear and DLinear.

\paragraph{Ablation study. } 
For ablation study, we made three key modifications and evaluated their impact.
We described the changes in Table \ref{tb:ablation}. 
To begin with, the logarithmic binning showed minimal information loss when converting tokens to binned tokens, with less performance decline compared to uniform interval binning.
Second, adding global embedding significantly enhanced accuracy. 
Last, simplifying the feed-forward network improved performance, boosted calibration speed, and lowered energy consumption.

\paragraph{Calibration tradeoff. } 
We analyzed RMSE, FLOPS, memory usage, and model parameter size to determine whether the model effectively achieves the desired balance (see Figure~\ref{fig:exp_efficiency}).
Linear models are fast and lightweight but lack the accuracy needed for calibration tasks.
Transformers generally outperform linear models in accuracy but are less efficient. 
While Transformer and Informer have high computational demands (\textit{e.g.}, memory footprint and FLOPS), PatchTST, iTransformer, and \textsc{Tesla} balance efficiency and effectiveness, with \textsc{Tesla} achieving the highest accuracy without significant overhead across efficiency metrics.

\paragraph{Sequence length. }
To evaluate the impact of the sequence length,  
we compared the most complex linear- and Transformer-based methods with \textsc{Tesla} (see Figure \ref{fig:exp_window}).
In general, increasing sequence length generally improves accuracy, 
but excessively long sequences degrade performance (see Figure~\ref{fig:exp_window}(a)). 
Note that the improvement rate also differs. 
DLinear showed slight improvement, while Transformers exhibited significant accuracy gains.
\textsc{Tesla} maintains efficiency in FLOPS at less than twice that of DLinear, whereas Transformers are considerably less efficient for extended sequence lengths.
For this reason, we set \(N=360\) as our optimal length for all models in our experiment.

\begin{figure}[t]
\centering
\includegraphics[width=0.92\linewidth]{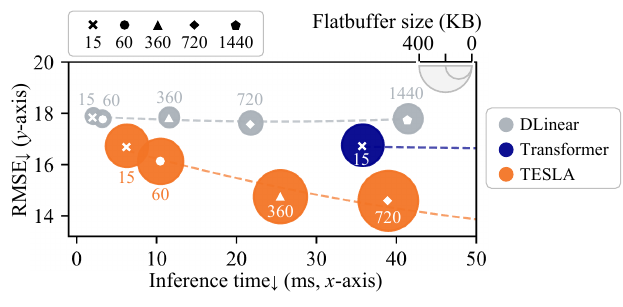}
  \caption{
  Evaluation of the Arduino Nano 33 BLE Sense microcontroller. The scatter map illustrates RMSE, inference time, and Flatbuffer size (diameter of the circles) with varying windows (15, 60, 360, 720, 1440) for average \(\textbf{PM}_\textbf{10}\) performance across three regions. Missing window sizes indicate non-applicability. The dotted line represents the trend.}
  \label{fig:exp_board}
\end{figure}

\paragraph{Case study. }
We visually compared the calibration results of the most practical linear-based model (DLinear), Transformer-based model (iTransformer), and \textsc{Tesla} (see Figure \ref{fig:exp_case}).
\textsc{Tesla} scored higher accuracy than iTransformer in most cases, with an average improvement of 3.81\%. 
Although this difference does not seem to significantly impact the calibration results,
\textsc{Tesla} achieves a remarkable 34.86\% improvement over iTransformer under \textbf{High-distribution} scenarios.
Note that the gap is even more notable when compared to DLinear. 

\paragraph{Performance on microcontroller. } 
To evaluate performance in an IoT embedded environment, we measured applicability and inference speed on a microcontroller using representative models (see Figure \ref{fig:exp_board}).
DLinear is small-sized and supports long sequences but shows low accuracy.
In contrast, Transformer handles only short sequences due to high computational demands.
\textsc{Tesla} has a model size similar to Transformer but requires far less computation.

\section{Discussions and Limitations}

We discuss the implications and limitations of our work.

\paragraph{Ineffective linear models. }
Newly-crafted linear models underperformed compared to Transformer-based models (Table \ref{tb:result_main}), contrary to trends in forecasting tasks. 
This discrepancy stems from differences in dataset scale: calibration tasks involve larger datasets, favoring complex architectures with more parameters. 
Furthermore, the minimal accuracy gap between Linear and DLinear suggests that simple averaging is ineffective for calibration. 
These results highlight that while effective for forecasting, the linear approach may not be suitable for calibration tasks.

\subsubsection{Effectiveness of key modification in \textsc{Tesla}.}
Table \ref{tb:ablation} demonstrates the effectiveness of \textsc{Tesla}’s three key modifications: 
(i) Building upon PatchTST’s uniform patching approach, \textsc{Tesla} introduces logarithmic binning, which mitigates performance degradation by emphasizing recent values and adapts better to calibration tasks;
(ii) Integrating global embedding, motivated by iTransformer, improves accuracy and extends its multivariate design to univariate scenarios; and 
(iii) Simplifying the feed-forward network to feature-wise linear, inspired by linear forecasters, highlights the importance of relationships between adjacent values and even achieves performance improvements in calibration tasks; 
In short, \textsc{Tesla} benefits from the strengths of both Transformers and linear modeling techniques.

\paragraph{\textsc{Tesla} leads in effectiveness and efficiency.}
\textsc{Tesla} is the \textit{desideratum} for addressing both efficiency and effectiveness in calibration tasks, balancing these aspects better than its counterparts (Figure \ref{fig:exp_efficiency}). 
While PatchTST and iTransformer also achieve this balance, \textsc{Tesla} stands out as the most accurate model with no significant overhead across all efficiency metrics, making it the ideal choice for maximizing performance under resource constraints.

\subsubsection{Mean trap phenomenon.}
The minimal gap between RMSE and MAE arises from the ``mean trap,'' where limited variation in time series data makes these metrics less sensitive to calibration differences. 
Even in high-distribution scenarios with significant calibration improvements (see Figure \ref{fig:exp_case}), these metrics often fail to capture the differences. 
This underscores the value of using a model like \textsc{Tesla}, where calibration improvements, though seemingly minor in metrics, can have significant real-world impacts.

\subsubsection{Static experimental settings.}
Although we adopt standard settings used in calibration studies \cite{ahn24sendal, narayana24sensbert} and \textsc{Tesla} has demonstrated adaptability to IoT environments, the static nature of this approach requires recalibration or drift correction.
This limitation can be alleviated as larger datasets become available.
We further discuss this in Appendix \ref{sec:appendix_limitation}.


\section{Conclusion}
\label{sec:07-conclusion}

This study introduced \textsc{Tesla}, a Transformer-based calibration model for low-cost sensors in practical IoT systems. 
We first identified three challenges that must be addressed to adapt these models for calibration tasks. 
Then, \textsc{Tesla} addresses the challenges in the following manner:
(i) Mitigate the attention bottleneck by employing logarithmic binning to reduce the number of tokens, 
(ii) Integrate multi-view embeddings, and 
(iii) Ensure fast processing by replacing feed-forward operations with linear layers. 
Results demonstrated that \textsc{Tesla} excelled in effectiveness and efficiency.


\appendix
\renewcommand{\thefigure}{A\arabic{figure}} 
\renewcommand{\thetable}{A\arabic{table}} 
\setcounter{figure}{0}
\setcounter{table}{0}

\section{Appendices}
\subsection{Dataset details \label{sec:appendix_dataset}}
This section outlines the data characteristics and the criteria used to select subsets for the calibration task. 
As discussed in Section \ref{sec:03_prelim}, applying benchmarks for time series forecasting to calibration poses several challenges.
The SensEURCity dataset \cite{van2023senseurcity}, specifically designed for calibration tasks, provides a robust foundation for addressing these challenges.

To meet task-specific requirements, we select subsets that reflect diverse distributions of measurements for training to capture general measurement patterns. 
Specifically, we carefully selected: 
(i) data collected during the first collocation period for each region, as this period provides sensor readings from the same location and time, ensuring consistent calibration under uniform environmental conditions; and
(ii) defining Plantower PMS5003 as a low-cost sensor and Alphasense OPC-N3 as a reference reading, categorized by price.
The metadata for these subsets is shown in Table \ref{tb:metadata}.

To maintain data quality, we excluded removed outliers or invalid values, and even discarded any sensor for which more than half of its readings were removed. 
In practical IoT scenarios, sensor data is collected in real-time, so no further refinement is applied beyond removing invalid data.

\begin{table}[t]

\centering
\setlength\tabcolsep{8pt}
\resizebox{0.95\columnwidth}{!}{%
\begin{tabular}{lcccc}
    \toprule
    \textbf{Region} & \textbf{Antwerp (Ant.)} & \textbf{Oslo} & \textbf{Zagreb (Zag.)} \\
    \midrule
    \#Sensors & 24 & 21 & 14 \\
    Data format & 3,113,955\(\times\)6 & 810,314\(\times\)6 & 939,838\(\times\)6 \\
    Start time & 20.04.03 16:12 & 20.09.17 09:56 & 20.05.22 16:06 \\
    End time & 20.06.15 07:07 & 20.10.14 08:56 & 20.07.08 12:00 \\
    Granularity & 1min & 1min & 1min \\
    \bottomrule
\end{tabular}

}
\caption{Metadata for datasets used in our experiment with the data format (\#data for all sensors, \#features). 
Each region includes features \(\textbf{PM}_\textbf{10}\), \(\textbf{PM}_\textbf{2.5}\), and \(\textbf{PM}_\textbf{1}\) from both low-cost and reference sensors.}
\label{tb:metadata}
\end{table}

\subsection{Detailed evaluation setup \label{sec:appendix_setup}}
This section describes the evaluation setup for the calibration task and the reasoning behind its configuration. 
Unlike forecasting tasks, where models predict future values based on past time series, the calibration task focuses on learning general patterns from the temporal behavior of each sensor to ensure robust performance across diverse conditions. 
To align with the objectives of calibration, our method modifies the train-validation-test split to focus on capturing general patterns from sensor data rather than temporal prediction.

In the SensEURCity dataset, each sensor is identified by its unique ID and organized in alphabetical order within each region. 
The second-to-last sensor is used as the validation set, the last sensor as the test set, and all remaining sensors are used for training. 
This configuration ensures that the model learns patterns from diverse training sensors while maintaining consistent evaluation across regions.

\subsection{Baselines details \label{sec:appendix_baseline}}
This section introduces a detailed description of the baselines used in our experiment: (1) Linear-based models, and (2) Transformer-based models. 
Recent studies suggest that LSTM-based models are less commonly used in time series forecasting, as recent advancements have shifted focus to other architectures.
Therefore, our experiment follows commonly used configurations found in recent research \cite{zeng2023dlinear,nie2023timeseriesworth64,liu2024itransformer} by adopting linear-based and Transformer-based models.
The detailed descriptions of each baseline are as follows:

\begin{itemize}[leftmargin=1em]

\item\textbf{Linear} \cite{freedman2009statistical}: Linear straightforwardly uses weighted inputs, achieving moderate performance in time series prediction tasks.
\item\textbf{NLinear} \cite{zeng2023dlinear}: NLinear involves normalizing the input time series data to its last value, which focuses the analysis on detailed dynamics.
\item\textbf{DLinear} \cite{zeng2023dlinear}: DLinear explicitly separates the time series into trends and residuals and uses individual linear layers.
\item\textbf{Transformer} \cite{vaswani2017attentionneed}: Transformer uses a self-attention mechanism to weigh input data effectively, but it suffers from a quadratic computational bottleneck.
\item\textbf{Informer} \cite{zhou2021informer}: Informer reduces the bottleneck by using selective queries, marking the first attempt to apply a Transformer architecture to a prediction task.
\item\textbf{PatchTST} \cite{nie2023timeseriesworth64}: PatchTST enhances both effectiveness and efficiency by employing uniformly-sized patches as tokens, instead of using single values as tokens. 
\item\textbf{iTransformer} \cite{liu2024itransformer}: iTransformer extends the patch scope of PatchTST to its maximum, treating each sensor as an individual token. 
\end{itemize}

\subsubsection{Comparison to \textsc{Tesla}. } 

\textsc{Tesla} combines the advantages of various time series forecasters for calibration tasks. 
First, unlike PatchTST, it emphasizes recent data more effectively by binning on a logarithmic scale. 
Additionally, while iTransformer is only effective with multiple sensors or multivariate time series, \textsc{Tesla} adapts to both local and global contexts, making it suitable for univariate applications. 
Finally, we adapted the token-wise feedforward neural network to a feature-wise linear layer to enhance the calibration speed comparable to that of linear forecasters.

\subsection{More discussions \label{sec:appendix_limitation}}
\textsc{Tesla} achieves fast inference speed and high calibration accuracy, but the limited availability of large-scale calibration datasets constrains its effectiveness.
First, additional experiments using varied sensor data, such as temperature and humidity, are required to evaluate \textsc{Tesla}’s performance in more diverse scenarios.
Second, the reliability of the reference readings remains a concern, as the high-cost sensor used in this study may not provide sufficient precision. Future research should explore using more advanced sensors or alternative references to enhance calibration accuracy.
Lastly, using pre-trained weights requires periodic updates through server communication or firmware updates to sustain consistent performance, emphasizing the importance of developing dynamic training methods in future studies.

\section*{Acknowledgements}
We thank Miro Co., Ltd. for their valuable help. This work was supported in part by the National Research Foundation of Korea (NRF) grant funded by the Korea government (MSIT) (No. NRF-2022R1C1C1012408), in part by Institute of Information \& communications Technology Planning \& Evaluation (IITP) grants funded by the Korea government (MSIT) (No. 2022-0-00448/RS-2022-II220448, Deep Total Recall: Continual Learning for Human-Like Recall of Artificial Neural Networks, and No. RS-2022-00155915, Artificial Intelligence Convergence Innovation Human Resources Development (Inha University)), and in part by INHA UNIVERSITY Research Grant.

\bibliography{aaai25}

\end{document}